\def\year{2021}\relax
\documentclass[letterpaper]{article} 
\usepackage{aaai21}  
\usepackage{times}  
\usepackage{helvet} 
\usepackage{courier}  
\usepackage[hyphens]{url}  
\usepackage{graphicx} 
\usepackage{natbib}  
\usepackage{caption} 
\nocopyright{}

\title{Neural Model-based Optimization with Right-Censored Observations}
\author {
    Katharina Eggensperger,\textsuperscript{\rm 1}
    Kai Haase,\textsuperscript{\rm 1}
    Philipp M\"uller,\textsuperscript{\rm 1}
    Marius Lindauer,\textsuperscript{\rm 2}
    Frank Hutter\textsuperscript{\rm 1,3}\\
}
\affiliations {
    \textsuperscript{\rm 1} University of Freiburg, Germany \\
    \textsuperscript{\rm 2} Leibniz University Hannover, Germany \\
    \textsuperscript{\rm 3} Bosch Center for Artificial Intelligence \\
    eggenspk@cs.uni-freiburg.de, haasek@informatik.uni-freiburg.de, muelleph@informatik.uni-freiburg.de, lindauer@tnt.uni-hannover.de, fh@cs.uni-freiburg.de
}

\usepackage{latexsym}
\usepackage{booktabs}       
\usepackage{amsmath,amssymb,amsfonts,dsfont,amsthm}  
\usepackage{bm} 
\usepackage[dvinames,dvipsnames]{xcolor}
\usepackage{pifont}
\usepackage{comment}
\usepackage{enumitem}
\usepackage{url}
\theoremstyle{plain} 

\usepackage{multirow}
\usepackage{subcaption}
\usepackage{caption}
\usepackage[utf8]{inputenc}
\usepackage{pdfpages}

\definecolor{mred}{HTML}{E41A1C}
\definecolor{mblue}{HTML}{377AB8}
\definecolor{mpalered}{HTML}{FB9A99}

\usepackage{pgfplots}
\usepackage{tikz}
\usetikzlibrary{shapes.geometric}
\usetikzlibrary{positioning,shapes,shadows,arrows,calc}
\usetikzlibrary{intersections}
\pgfdeclarelayer{background}
\pgfdeclarelayer{foreground}
\pgfsetlayers{background,main,foreground}

\newcommand{\wrt}{w.r.t.~}

\newcommand{\pcs}[0]{\bm{\Lambda}}

\newcommand{\cost}[0]{c}

\newcommand{\ind}{I}
\newcommand{\real}{\mathbb{R}}

\newcommand{\conf}[0]{\bm{\lambda}}
\newcommand{\cutoff}[0]{\kappa}

\newcommand{\roar}{\textit{Rand}}

\newcommand{\saps}{\textit{Saps}}

\DeclareMathOperator*{\argmin}{arg\,min}

\begin{document}
\maketitle

\begin{abstract}
In many fields of study, we only observe lower bounds on the true response value of some experiments. When fitting a regression model to predict the distribution of the outcomes, we cannot simply drop these \emph{right-censored observations}, but need to properly model them. In this work, we focus on the concept of censored data in the light of model-based optimization where prematurely terminating evaluations (and thus generating right-censored data) is a key factor for efficiency, e.g., when searching for an algorithm configuration that minimizes runtime of the algorithm at hand. Neural networks (NNs) have been demonstrated to work well at the core of model-based optimization procedures and here we extend them to handle these censored observations. We propose (i)~a loss function based on the Tobit model to incorporate censored samples into training and (ii) use an ensemble of networks to model the posterior distribution. To nevertheless be efficient in terms of optimization-overhead, we propose to use Thompson sampling s.t. we only need to train a single NN in each iteration. Our experiments show that our trained regression models achieve a better predictive quality than several baselines and that our approach achieves new state-of-the-art performance for model-based optimization on two optimization problems: minimizing the solution time of a SAT solver and the time-to-accuracy of neural networks.
\end{abstract}

\section{Introduction}
\label{sec:intro}

When studying the outcome of an experiment we might only observe a lower or an upper bound on its true value -- a censored observation. Such censored data is present in many applications, in particular if individual observations are costly in terms of time and resources. For example when studying the impact of a fertilizer, the quantity of a toxin can drop below the detection level of the measurement device; when studying the expected lifetime of hard drives, the study might be stopped before all hard drives have exceeded their lifetime; or when studying the damage of insect pests to growing crops, some plants might still be healthy at harvesting time. 
To analyze such \emph{time-to-event} data it is crucial to handle censored data correctly to avoid over- or underestimation of the quantity of interest.

In this work, we study the concept of censored data in the light of model-based optimization where we are interested in minimizing an objective function describing the time required to achieve a desired outcome. We focus on optimization procedures that actively terminate evaluations to speed up the optimization by spending less time on poorly performing evaluations and thus generate right-censored data. 
This censoring mechanism enables to efficiently tune costly objectives and can speed up optimization by orders of magnitude~\cite{hutter-ecmltalk17a,kleinberg-ijcai17a,kleinberg-neurips19a,weisz-icml18anew,weisz-icml19anew}.
Censoring strategies thus substantially contribute to the state of the art for automated algorithm configuration; the potential of which has been demonstrated in many domains of AI by providing speedup factors of up to $3000\times$ in satisfiability solving~\citep{hutter-aij17a}, 118$\times$ in AI planning~\citep{vallati-socs13a}, $100\times$ for reinforcement learning~\citep{falkner-icml18anew}, 28$\times$ for time-tabling~\citep{chiarandini-patat08a}, $52\times$ for mixed integer programming~\citep{hutter-cpaior10a} and $14\times$ for answer set solving~\citep{gebser-lpnmr11a}.
However, the efficiency of the model-based optimizers stands and falls with the quality of the empirical performance model at their core. 

Bayesian optimization (BO)~\citep{shahriari-ieee16a}, as one of the most studied model-based approaches, commonly uses Gaussian processes (GPs)~\citep{mockus-jgo94,brochu-arXiv10a,snoek-nips12a} or random forests (RFs)~\citep{feurer-nips2015a,caceres-gecco17a,candelieri-jgo18a}, but recently methods based on neural networks (NNs) have been shown to perform superior on some applications~\citep{schilling-ecmlpkdd15a,snoek-icml15a,springenberg-neurips2016,perrone-neurips18a,white-metalearn19a}. 
In this work, we extended this NN-based Bayesian optimization to work in the presence of partially censored data. 
Specifically, our contributions are:
\begin{enumerate}
    \item We propose to use the Tobit loss for training NNs to properly handle noisy and censored observations.
    \item We study the impact of censoring on the training of NNs using this loss function.
    \item We show that ensembles of NNs, trained on censored data,  
    in combination with Thompson Sampling yield efficient optimization procedures.
    \item We demonstrate the benefit of our model on two runtime minimization problems, outperforming the previous state of the art.
\end{enumerate}

\section{Formal Problem Setting}
\label{sec:problemsetting}

First, we formally describe the problem setting we address in this work by discussing the challenges of regression under censored observations and showing how these relate to model-based optimization.

\subsection{Regression under Censored Data}
\label{ssec:censreg}

We focus here on a specific type of data where the variable of interest is the time\footnote{We use the term \emph{time} for simplicity, but we note that this notation also generalizes to other metrics, e.g. CPU cycles, gradient descent steps, training epochs, and number of memory accesses.} until an event occurs, e.g., until an algorithm has solved a task at hand. If this event does not occur within the time allocated for an evaluation, we obtain a \emph{right-censored observation}, i.e. a lower bound on its actual runtime, but not the actual value. There exist two other types of censored values which we do not consider here since they do not occur during the type of optimization we are interested in: left-censored values for which the start time is unknown, but the event has happened (e.g., we do not know when the algorithm has started) and interval-censored values for which we know that the event has happened within some time interval (e.g., we know that the algorithm has solved the problem within $10$ to $20$ seconds). 

Our goal is to build a regression model trained on data $\mathcal{D}=(\bm{x}_i, y_i, \ind_i)_{i=1}^n$, where $\bm{x} \in \real^d$ is the $d$-dimensional input, $y \in \real$ is the observed value and $\ind \in \{1, 0\}$ is a binary variable indicating whether the value of this observation has been censored or not; if $\ind_i = 1$, the $i$-th observation has been censored and $y_i$ is only a lower bound of the true value. Furthermore, we assume that the true (non-censored) value $y_i$ is generated by a stochastic process as is the case for, e.g., observing the time of a non-deterministic algorithm. We will explain in the next section how such data is generated.
\begin{figure}[t]
    \includegraphics[height=10.5em]{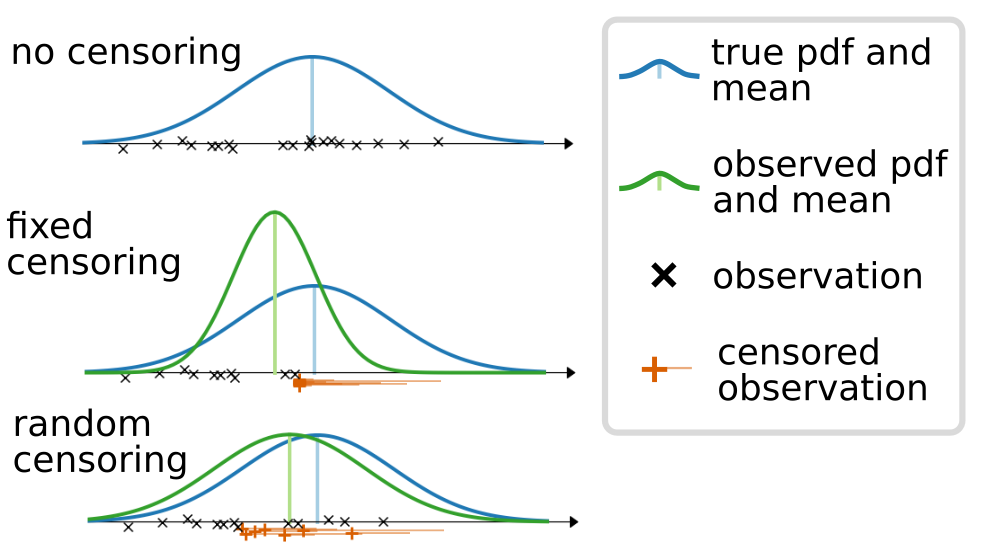}
    \caption{True and observed normal distributions with and without censored observations.}
    \label{fig:censdist}
\end{figure}
In Figure~\ref{fig:censdist}, we visualize how the observed distribution changes when some of the samples are censored. The upper plot shows the true distribution which generated the observations, and the middle (lower) plot shows how censoring at fixed (randomly chosen) censoring thresholds skew the empirically observed distribution. Fitting a maximum-likelihood model on such data could lead to substantial underestimations of the target value's true mean.

\subsection{Sequential Model-based Optimization Using Censoring Strategies}
%
\begin{figure}
    \scalebox{0.76}{    \centering
    \tikzstyle{activity}=[rectangle, draw=black, rounded corners, text centered, text width=7em, fill=white, drop shadow]
\tikzstyle{data}=[rectangle, draw=black, text centered, fill=black!10, text width=4em, drop shadow]
\tikzstyle{myarrow}=[->, thick]
\begin{tikzpicture}[node distance=10em]

	\node (model) [activity] {Fit probabilistic model $\mathcal{M}$ on $\mathcal{D}$};
	\node (select) [activity, right of=model, node distance=4.4cm] {Select $\conf$ based on $\mathcal{M}$};
	
	\node (racing) [activity, below of=model, node distance=1.5cm, xshift=2.2cm] {Racing \& \\ Adaptive Capping};
	\node (target) [activity, below of=racing, node distance=2.0cm] {Target\\ Algorithm};

	\draw[myarrow] (model) -- (select);
	\draw[myarrow] (select) |- ($(racing.east)+(0.5,-0.6)$);
	\draw[myarrow] ($(racing.west)+(-0.5,-0.6)$) -| node [left, yshift=0.8cm] {$\mathcal{D}=\langle(\conf_i, c_i, \ind_i) \rangle_i$} (model);
	\draw[myarrow] ($(racing.west)+(-0.5,-0.6)$) -| node [left, yshift=0.4cm] {incumbent $\conf_\text{inc}$} (model);

    \path ($(racing.south)+(-0.2,-0.0)$) edge[bend right, ->, thick] node [left] {$\conf, \cutoff$} ($(target.north)+(-0.2,0)$);
	
	\path ($(target.north)+(0.2,0)$) edge[bend right, ->, thick] node [right] {$c$} ($(racing.south)+(0.2,-0.0)$) ;
	
	\begin{pgfonlayer}{background}

    	\path (racing -| racing.west)+(-0.5,0.9) node (resUL) {};
    	\path (target.east |- target.south)+(0.5,-0.2) node(resBR) {};
    	\path [rounded corners, draw=black!80, dashed] (resUL) rectangle (resBR);
    	
    \end{pgfonlayer}
\end{tikzpicture}}
    \caption{Model-based algorithm configuration using racing and censoring strategies.}
    \label{fig:mbac}
\end{figure}
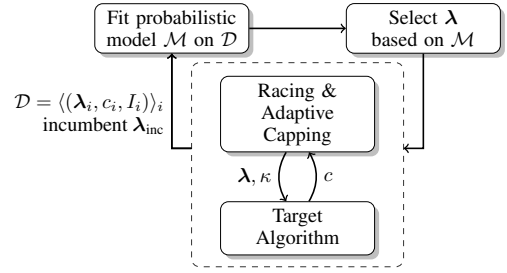
We aim to find a solution $\conf^*$ minimizing a black-box function describing the stochastic cost metric $\cost: \conf \rightarrow{} \real^+$ of evaluating $\conf \in \pcs$ (e.g., the time to obtain a solution):

\begin{equation}
    \conf^* \in \argmin_{\conf \in \pcs} \mathbb{E}[\cost(\conf)].
\end{equation}
Sequential model-based optimization, especially Bayesian Optimization (BO;~\cite{brochu-arXiv10a,shahriari-ieee16a}), has been demonstrated to work well for such problems. This optimization procedure typically iterates over three phases: (i) fitting a probabilistic model on all observations, (ii) choosing the next configuration to evaluate based on the model and an acquisition function, e.g., Expected improvement~\citep{jones-jgo98a} and (iii) evaluating this configuration. In this work, we focus on how to conduct step (i) in the presence of censored observations.

As discussed before, we consider optimization procedures that actively cap the evaluation of poorly performing configurations. A well known example for such procedures is algorithm configuration (AC)~\citep{hutter-jair09a} based on BO methods using a censoring strategy. In Figure~\ref{fig:mbac}, we provide an overview of this iterative procedure. AC in principle performs the same steps as BO, but instead of evaluating a selected configuration in isolation, it is raced against the best configuration seen so far and might be prematurely terminated if it performs worse. This racing procedure accommodates for noise across repeated evaluations of the same configuration using different seeds or solving different problem instances.
Thus, besides choosing a configuration $\conf$ to evaluate next, AC methods in addition also define a cutoff time $\cutoff$ at which each evaluation will be terminated. The cutoff $\cutoff$ can either be a globally set cutoff $\cutoff_{max}$ or be adapted for each new run, such that new configurations use at most as much time as the best configuration seen so far~\cite{hutter-jair09a}. For a  configuration $\conf_i$ at the $i$-th observation, we then observe 

\begin{equation}
\cost_i = \min\left(\cutoff_i, \cost(\conf_i)\right)
\end{equation}
with $\cost(\conf_i)$ being a sample from the cost metric (which we only fully observe if $\cost(\conf_i) \le \cutoff_i$). These observations form the training data $\mathcal{D} = (\conf_i, \cost_i, \ind_i)_{i=1}^n$ for the model (i.e. $(\bm{x}_i, y_i, \ind_i)_{i=1}^n$ as we discussed before) to be used in BO.
AC has been used with random forests (RFs) to guide the search procedure and the resulting model-based methods define one direction of state-of-the-art AC methods~\citep{hutter-lion11a,ansotegui-ijcai15a,caceres-gecco17a}.

\section{Related Work on Modeling under Censored Data}
\label{sec:relwork}
Handling censored data has a long history in the field of survival analysis~\citep{kleinbaum-survivalanalysis10a,haider-jmlr20a} studying statistical procedures to describe survival or hazard functions. Commonly used methods are the Kaplan-Meier estimator~\citep{kaplan-58a}, the Cox Proportional Hazards model~\citep{cox-72a}, the accelerated failure time model~\citep{kalbfleisch-accelerated11a} and also extensions to regression models, such as NNs~\cite{katzman-bmc18a,lee-aaai18a} and RFs~\citep{ishwaran-applstats09a}. However, here, we are not interested in the impact of different risk factors or the risk of the occurrence of the event at a specific time step (the so-called hazard function), but we are interested in the mean survival time given a configuration. Additionally, for methods modeling the survival function relying on a non-parametric survival function, such as the Kaplan-Meier estimator, the mean survival time is not well defined if the largest observation is censored, which is the case for model-based optimization.

Survival analysis methods in the context of modeling algorithm performance have been studied by~\citet{gagliolo-thesis10a} in order to construct online portfolios of algorithms to solve a sequence of problem instances. However, much more related to our work is the method introduced by \citet{schmee-techno79} for regression with censored data. This method iteratively increases the values of censored data using a two step procedure: (1) Fit a model to the current data and (2)~update the censored values according to the mean of the predicted normal distribution truncated at the observed censored value (to ensure that the updated value can never become smaller than the observed value). Steps (1) and (2) are repeated until the method converged or a manually defined number of iterations was performed. This procedure has been adapted to work with Random Forests and has been demonstrated to work well as a model for model-based optimization to minimize the runtime of stochastic algorithms~\citep{hutter-bayesopt11}. As we show in our experiments, this method performs similarly to our approach, but because of its iterative nature, it is slower than ours by at least a factor of $k$ if using $k$ iterations.

Other related work includes regression models for handling censored data in a specific context different from ours, e.g., a weighted loss function for fitting a linear regression for high-dimensional biomedical data~\citep{li-sdm16a} or a two-step procedure for Gaussian process regression to build a model that resembles the output of a (physical) experiment where observations might be censored due to unknown resource constraints~\citep{chen-arxiv19a}.

\section{Training a Neural Network on Censored Observations}
\label{sec:modeldesign}
%
\begin{figure*}[t]
    \centering
    \begin{tabular}{ccccc}
    {~~~~(a) NN (Ignore)} & (b) NN S\&H & (c) NN Tobit loss & (d) NN Ens. Tobit loss & (e) NN Ens. Tobit loss\\
\includegraphics[height=3cm,trim=0cm 0cm 2.3cm 0cm,clip]{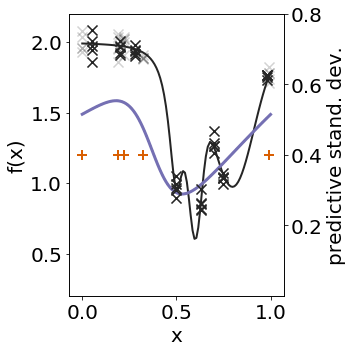} &
\includegraphics[height=3cm,trim=2.3cm 0cm 2.3cm 0cm,clip]{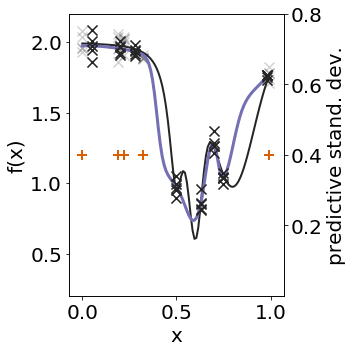} &
\includegraphics[height=3cm,trim=2.3cm 0cm 2.3cm 0cm,clip]{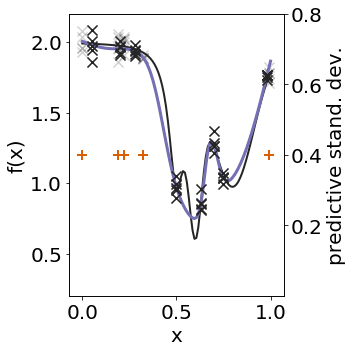} &
\includegraphics[height=3cm,trim=2.3cm 0cm 2.3cm 0cm,clip]{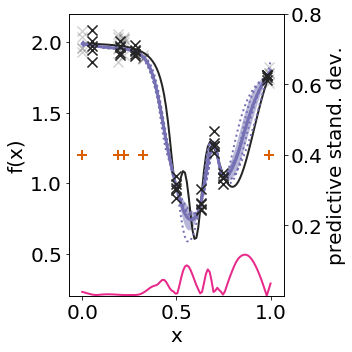} &
\includegraphics[height=3cm,trim=2.3cm 0cm 0cm 0cm,clip]{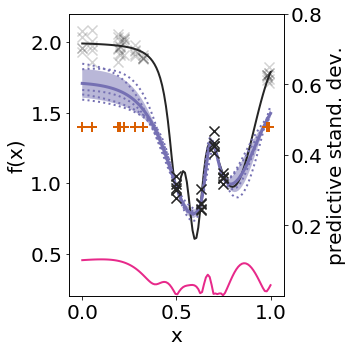} \\ 
\multicolumn{5}{c}{\includegraphics[width=0.9\textwidth]{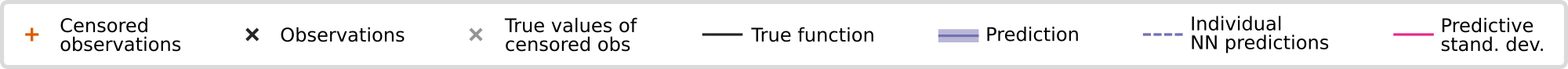}} \\
    \end{tabular}
    \caption{Comparing the predictive quality of NNs ignoring censoring information (a) using S\&H (b) using the Tobit Loss (c) and ensembles of NNs using the Tobit loss (d). We show results for random censoring in (a,b,c,d) and fixed censoring in (e).
    }
    \label{fig:visualcomp}
\end{figure*}
The model at the heart of state-of-the-art model-based optimization methods has to handle various challenges such as highly varying noise and most importantly for us: censored data. Neural Networks (NNs) provide a flexible learning framework based on (stochastic) gradient descent allowing to use any differentiable loss function. We will make use of this flexibility to incorporate censored data during training.

First, we want to model the value of interest as a random variable with a normal distribution, e.g. data fed into the model describing the cost of a stochastic algorithms can contain aleatoric noise. We model this parametric distribution by adding a second output neuron to the last layer modeling the variance~\citep{nix-icnn94}, i.e. our NN predicts a mean $\mu_i$ and variance $\sigma_i^2$ depending on the input $\conf_i$. To train our network on data $\mathcal{D} = (\conf_i, \cost_i, \ind_i)_{i=1}^n$, we use the negative log-likelihood (NLL), customary in deep learning:

\begin{equation}
\begin{aligned}
-\log\mathcal{L}\left((\hat{\mu}_{i}, \hat{\sigma}^2_{i})_{i=1}^n \mid \mathcal{D}\right) &= -\sum^n_{i=1} \log \phi(Z_i), \\
Z_i &= \frac{\cost_i-\hat{\mu}_{i}}{\hat{\sigma}^2_{i}},
\end{aligned}
\label{eq:nllh}
\end{equation}
with $\hat{\mu}_{i}$ and $\hat{\sigma}^2_{i}$ being the predicted mean and variance of the objective value for a configuration $\conf_i$ and $\phi$ being the standard normal probability density function.

Using this loss also for censored observations yields suboptimal estimators underestimating the true distribution (see Plot (a) in Figure~\ref{fig:visualcomp}). A simple solution would be to employ the iterative model-agnostic procedure proposed by \citet{schmee-techno79} (S\&H) to restore censored values before training the final model. However, this multiplies the training time depending on an additional hyperparameter, the number of iterations (e.g. for $5$ iterations we would need to train $5$ models). Furthermore, treating the censored information would be decoupled from the training procedure potentially misleading the estimator (see Plot (b) in Figure~\ref{fig:visualcomp}).

Instead, our approach relies on a solution handling censored data directly within the loss function. In the sum of likelihoods from Eq.~\eqref{eq:nllh} we correct the terms for censored observations. For right-censored data, all we know is a lower bound on the actual time of occurrence and this quantity can be described by $1-\Phi(Z_i)$, with $\Phi$ being the standard normal cumulative distribution function~\citep{gagliolo-cp06a,klein-survival06a}, also known as the Tobit likelihood function~\cite{tobin-econometria58}. These two terms complement each other by describing the information obtained from censored observations, i.e., the probability that the true value for this observation lies above the observed value, and the information obtained from uncensored observations, i.e., the probability of observing the observed value. This yields the following loss function we use to train our network on data $\mathcal{D} = (\conf_i, \cost_i, \ind_i)_{i=1}^n$:

\footnotesize
\begin{equation}
\begin{aligned}
-\log\mathcal{L}&\left((\hat{\mu}_{i}, \hat{\sigma}^2_{i})_{i=1}^n \mid \mathcal{D}\right) = \\
-&\sum^n_{i=1} \log \left(\phi(Z_i)^{1-I_i}(1-\Phi(Z_i))^{I_i} \right), \\
    I_i &= 
    \begin{cases}
    0, & \text{if } \cost_i \le \cutoff_i \\
    1, & \text{otherwise}
    \end{cases}
\end{aligned}
\label{eq:tobitloss}
\end{equation}
\normalsize
Using this loss function allows to directly handle the information obtained from censored observations during training without any additional overhead as in S\&H and can yield predictions close to the actual values (see Plot (c) in Figure~\ref{fig:visualcomp}) predicting values for censored points to be higher than the observed censored value (orange markers).

\section{Integration into Model-Based Optimization}
\label{sec:integration}

Next, we will describe how we make use of our NN model within model-based optimization and how we use model uncertainty to select the next configuration to evaluate.

First, since the runtime distributions of stochastic algorithms are known to exhibit heavy-tails for each configuration~\citep{gomes-aaai97a,eggensperger-ijcai18a} and are constrained to be larger than zero we model the log of the observed runtime. Thus, we can model the distribution of its log-function values as a normal distribution. Such a parametric distribution assumption allows us to efficiently model the distribution of the quantity of interest~\citep{pintilie-competingrisks06a}.

Secondly, we note that the model uncertainty is different from the aleatoric noise modeled by the second output neuron (which is not relevant for modeling the mean performance of a configuration). A full Bayesian treatment of all weights in the NN would provide model uncertainty~\citep{neal-bayes96a}; however, this would come with a large computational overhead and additional hyperparameters to tune. Recently, \citet{lakshminarayanan-neurips17a} showed that an ensemble of NNs with random initializations also yield robust and meaningful predictive uncertainty estimates. We use this approach to model the posterior distribution of the objective function with the modification that we do not use the estimated aleatoric noise, since the goal in BO is to compute a predictive distribution over the \emph{true function value} at a point, not for the noisy observations at the point. Hence, we compute the mean $\mu_{\conf}$ and variance $\sigma^2_{\conf}$ for a configuration $\conf$ based on an ensemble of size $M$ as follows:

\begin{equation}
    \mu_{\conf} = \frac{1}{M} \sum_{m=1}^M\hat{\mu}_{\conf}^{(m)} \texttt{~~} \sigma^2_{\conf} = \frac{1}{M} \sum_{m=1}^M(\hat{\mu}_{\conf}^{(m)} - \mu_{\conf})^2.
\label{eq:meanvariance}
\end{equation}
Plot (d) in Figure~\ref{fig:visualcomp} shows the predictive mean and variance of an NN ensemble trained on noisy data with randomly censored data. The uncertainty favorably increases for unseen locations and the NNs were able to prune out the noise.

However, the most natural idea to simply use these ensembles comes with the risk of over-exploring poor regions: The mixed loss function we use to train our network only constrains the network to predict \emph{above} the censoring threshold and the models can, in principle, predict arbitrarily high values resulting in a potentially high empirical variance (see the left part of Plot (e) in Figure~\ref{fig:visualcomp}). This could make poorly performing areas, where we have only observed censored data, promising for acquisition functions using the predictive uncertainty despite the fact that none of the NNs predict the area to be promising. For this reason in practice we use \textit{Thompson Sampling}~\citep{thompson-biometrica33a}, i.e. we draw a sample function from the posterior distribution of the surrogate model and evaluate the configuration at its optimum. Thompson sampling based on ensembles has been shown to work well for online decision problems and reinforcement learning~\citep{osband-neurips16a,lu-neurips17a}. However, in contrast to this work, we do not maintain a set of ensemble members and draw from it to select the next configuration to evaluate but train a network from scratch in each iteration. This renders the requirement for reasonable uncertainty estimates in poor areas with only censored samples unnecessary and comes with the benefit of significantly reducing the computational requirement since we only need to train a single NN instead of the whole ensemble.
\begin{table*}[t]
    \small
    \centering
    \caption{RMSE values for ensembles of NNs ignoring censoring information (I), dropping censored observations (D), using 5 iterations of S\&H (S\&H), and using the Tobit loss (T); and under varying levels of randomized censoring (a higher threshold indicates less censoring, see text).}
    \begin{tabular}{c|
    r@{\hspace{2mm}}r@{\hspace{2mm}}r@{\hspace{2mm}}r|
    r@{\hspace{2mm}}r@{\hspace{2mm}}r@{\hspace{2mm}}r|
    r@{\hspace{2mm}}r@{\hspace{2mm}}r@{\hspace{2mm}}r|
    r@{\hspace{2mm}}r@{\hspace{2mm}}r@{\hspace{2mm}}r|}
 thresh & \multicolumn{4}{c}{10th percentile} & \multicolumn{4}{c}{20th percentile} & \multicolumn{4}{c}{40th percentile} & \multicolumn{4}{c}{80th percentile} \\
         Model & I & D  & S\&H & T & I & D  & S\&H & T & I & D  & S\&H & T & I & D  & S\&H & T \\
 \midrule
 branin & $62.6$      & $33.6$         & $31.5$         & $\mathbf{28.6}$ & $57.1$      & $\mathbf{21.4}$ & $22.7$   & $22.8$          & $53.3$      & $22.2$    & $22.4$   & $\mathbf{19.2}$ & $24.6$      & $15.1$         & $14.0$   & $\mathbf{8.2}$ \\
 camel  & $31.9$      & $25.5$         & $24.2$         & $\mathbf{23.6}$ & $31.4$      & $25.2$          & $23.9$   & $\mathbf{22.6}$ & $29.6$      & $25.8$    & $22.7$   & $\mathbf{20.4}$ & $15.3$      & $18.0$         & $11.7$   & $\mathbf{8.7}$ \\
 hart3  & $1.6$       & $\mathbf{0.2}$ & $0.3$          & $0.3$           & $1.2$       & $\mathbf{0.2}$  & $0.3$    & $0.2$           & $0.5$       & $0.2$     & $0.2$    & $\mathbf{0.2}$  & $0.2$       & $\mathbf{0.2}$ & $0.2$    & $0.2$          \\
 hart6  & $0.5$       & $0.2$          & $\mathbf{0.2}$ & $0.3$           & $0.3$       & $\mathbf{0.2}$  & $0.2$    & $0.2$           & $0.3$       & $0.2$     & $0.2$    & $\mathbf{0.2}$  & $0.2$       & $0.2$          & $0.2$    & $\mathbf{0.2}$ \\
    \end{tabular}
    \label{tab:rmsnllhcoinflip}
\end{table*}
%
\section{Studying the Impact of Censored Observations}
\label{sec:eval1}

Now, we turn to the empirical evaluation of our model.

\textbf{Implementation Details.}
Following~\citet{snoek-icml15a}, we use a fully connected feedforward NN with 3 hidden layers, each with $50$ neurons and $tanh$ activations. To train our network, we use SGD with momentum, a batch size of $16$, and a cyclic learning rate with one cycle increasing the learning rate to $1e^{-2}$ and then decreasing it again~\citep{smith-arxiv18a}. As regularization, we apply weight decay of $1e^{-4}$ and clip the gradients to be within $[-0.1, 0.1]$. To stabilize training, we use a softplus function for our second output neuron to ensure a positive value of the predicted standard deviation~\citep{lakshminarayanan-neurips17a} and initialize the bias for the second output neuron with the data deviation. As pre-processing, we normalize the training targets to have zero-mean and unit-variance and the input values to be in $[0,1]$. Our implementation uses PyTorch~\citep{paszke-neurips19a} and is integrated in the sequential model-based optimization framework SMAC~\citep{hutter-lion11a} using the Python3 re-implementation~\citep{SMACv3}. We will make it publicly available upon acceptance. All experiments were run on a compute cluster equipped with 2.80GHz Intel(R) Xeon(R) Gold CPUs.

\textbf{Baselines.} 
First, we study how well NNs trained with the Tobit loss function (T) perform on censored data. We compare our loss function to three alternative baselines using only NLL as a loss function: ignoring the censoring information, i.e., the ensemble trains on the same data but does not know which of the values are censored (I); dropping censored data, i.e., training the ensemble only on non-censored values (D); and using 5 iterations of the iterative procedure proposed by~\citet{schmee-techno79} (S\&H). For all methods, we trained ensembles of size $5$ and trained each network for $10\ 000$ epochs.

\textbf{Problem Setup.}
To study our NNs under controlled conditions, we consider $4$ synthetic global optimization problems: Branin (2D), Camelback (2D), and $2$ versions of the Hartmann Function (3D and 6D), to obtain training data. For each function $f$ we sample $100 \times D$ locations. We generated $10$ copies of our training data and added normally distributed noise with $\mu=0$ and \mbox{$\sigma=(\max(f)-\min(f)) \cdot 0.1$} to obtain noisy training data. Then, we use a threshold $\gamma$, e.g. the $20$th percentile of all observations, and censor all observations at a location $x_i$ for which $f(x_i) \ge \gamma$ with a probability increasing from $0$ to $1$. This mimics the data obtained during optimization, where censoring caps poorly-performing runs. To study the impact of censored samples, we study $4$ different thresholds ($10$th, $20$th, $40$th and $80$th) covering aggressive censoring and almost no censoring.

\textbf{Results.}
In Table~\ref{tab:rmsnllhcoinflip} we report the average \emph{root-mean-squared-error} (RMSE) using a $5$-fold CV of the ensemble mean prediction \wrt to the true function values. Looking at the overall results, T achieved the best results, not surprisingly, followed by S\&H since both methods incorporate censoring information. However, we note that S\&H requires $5\times$ the training time of using the Tobit loss. The alternative, I, performs worse since the model learns from biased data. In contrast, D can yield relatively good results for some cases if the uncensored samples provide enough information. Also, naturally all methods achieved the best results on the datasets with the least fraction of censored observations.

\begin{figure}
    \centering
        \begin{tabular}{@{\hspace{1mm}}c@{\hspace{1mm}}c@{\hspace{1mm}}}
    \includegraphics[height=2.2cm,trim=0.5cm 0cm 0cm 0cm,clip]{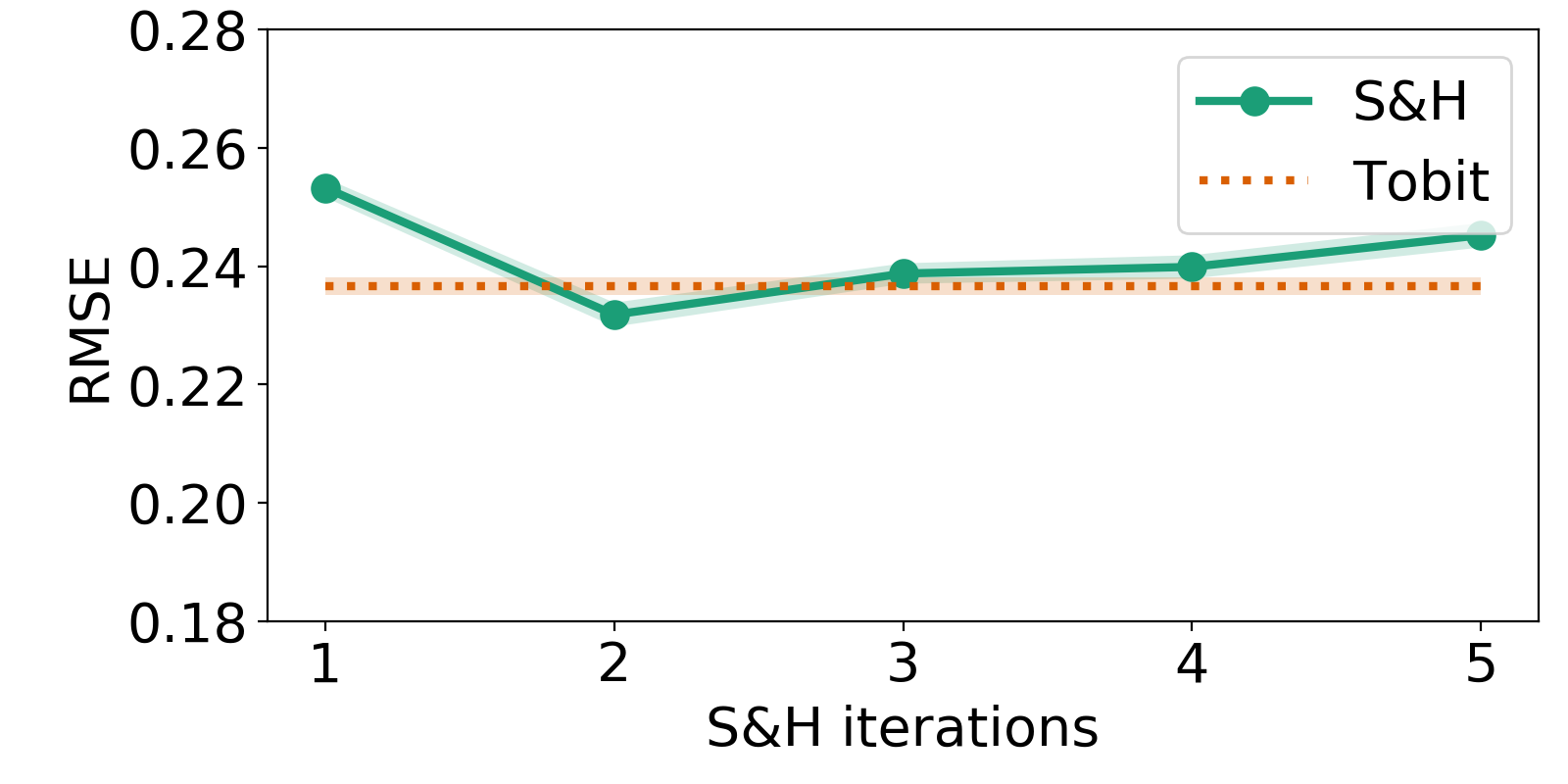} &
    \includegraphics[height=2.2cm,trim=3.25cm 0cm 0cm 0cm,clip]{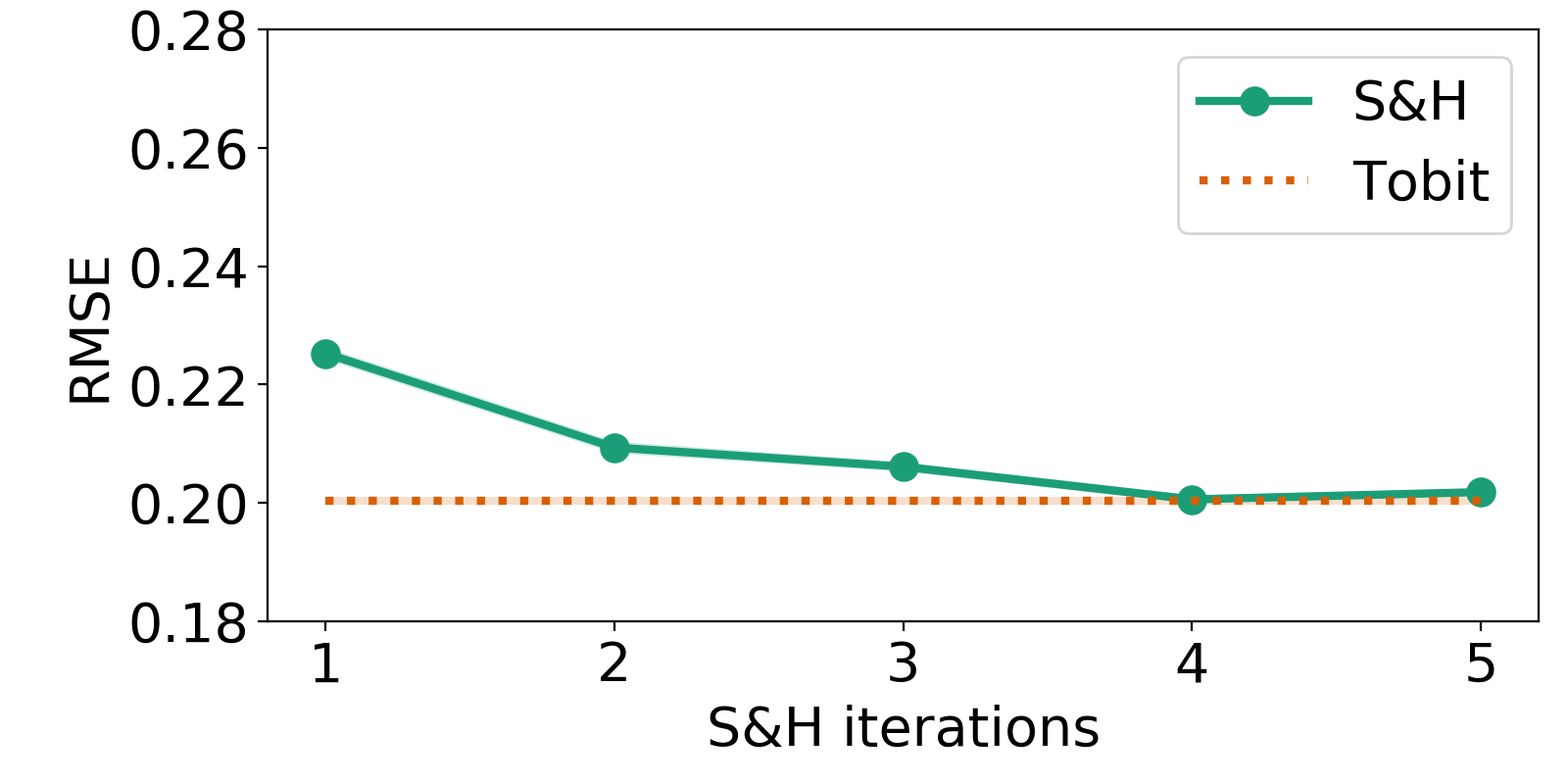} \\
        \end{tabular}
    \caption{Mean and variance of RMSE across a $5$-fold CV of an ensemble built by using $5$ iterations of the S\&H algorithm and using the Tobit loss. Left: Hartmann6 with aggressive censoring starting at the $20$th percentile and $52$\% censored data. Right: Hartmann6 with mild censoring starting at the $80$th percentile and $19.5$\% censored data.}
    \label{fig:shnet}
\end{figure}

Furthermore, we take a closer look at using S\&H for two exemplary scenarios in Figure~\ref{fig:shnet} and show how the RMSE changes with each iteration. Obviously, the largest improvement happens in the beginning and might improve or converge with further iterations. However, it is unclear beforehand how many iterations are required to achieve the best results, making S\&H much harder to use in practice.

\begin{figure*}[t]
    \centering
    \begin{tabular}{c@{\hspace{2mm}}cc@{\hspace{2mm}}c}
    \multicolumn{2}{c}{$QCP_{q095}$} & \multicolumn{2}{c}{$Adult$}  \\
    \includegraphics[width=0.22\textwidth]{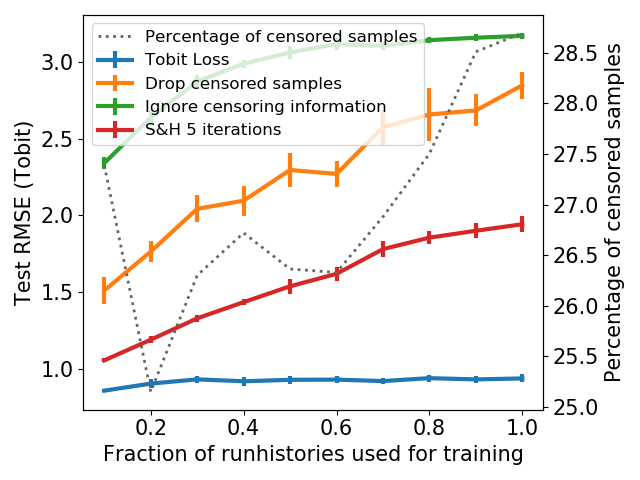} &
    \includegraphics[width=0.22\textwidth]{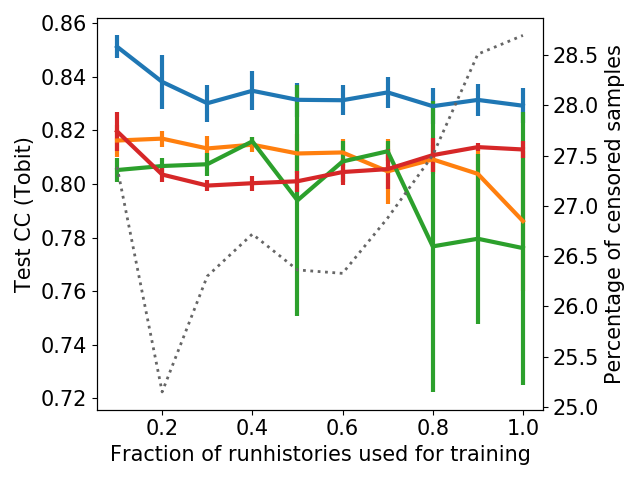} &
    \includegraphics[width=0.22\textwidth]{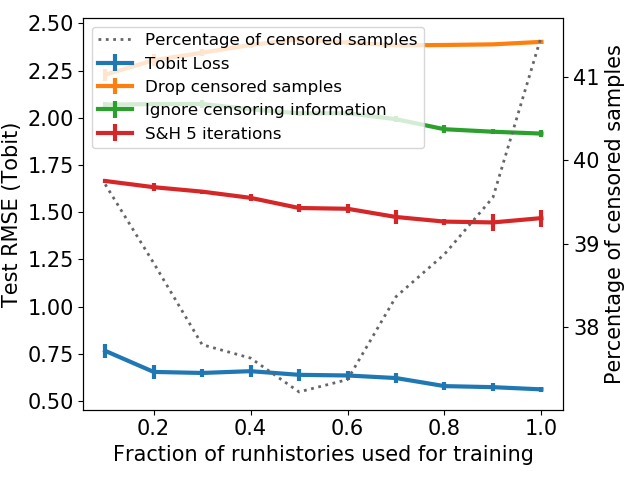} &
    \includegraphics[width=0.22\textwidth]{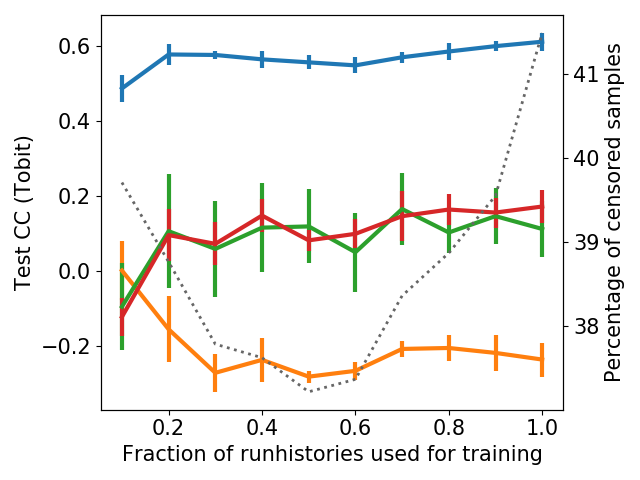} \\
    \midrule
    Ignore censoring & Drop censored data & S\&H $5$ iterations & Tobit Model \\
    \includegraphics[width=0.2\textwidth]{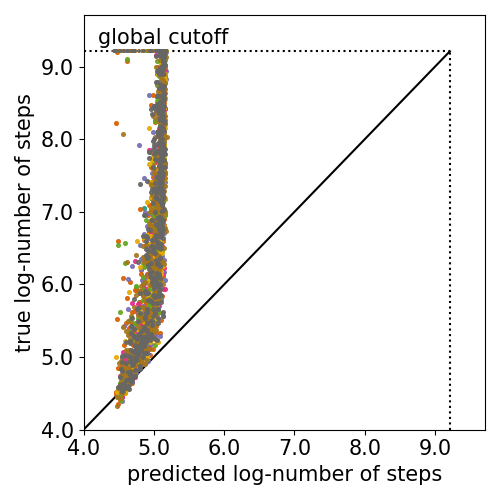} &
    \includegraphics[width=0.2\textwidth]{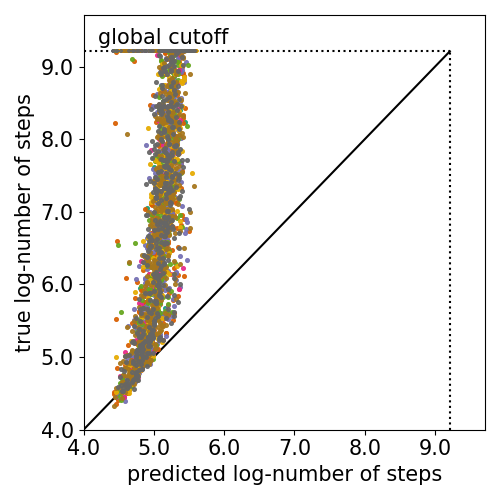} &
    \includegraphics[width=0.2\textwidth]{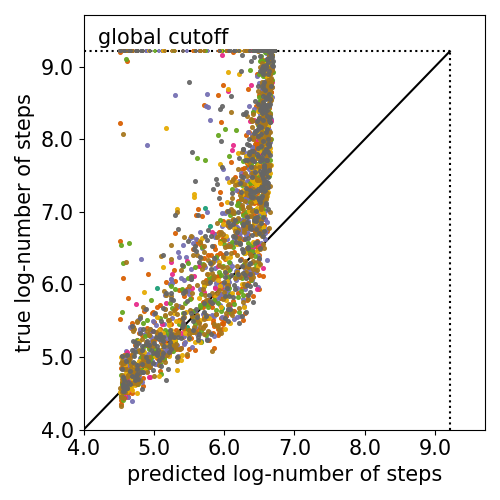} & 
    \includegraphics[width=0.2\textwidth]{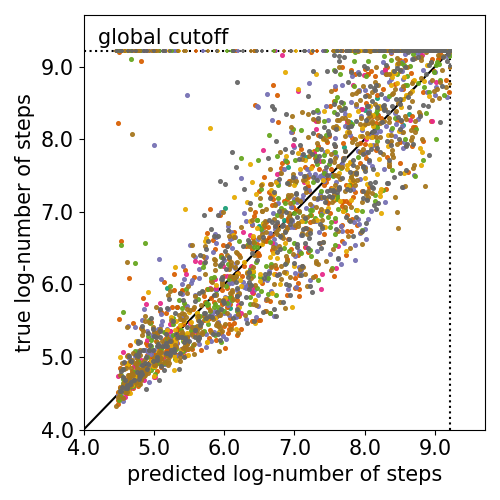}\\
    \hspace{0.2cm}\\
    \end{tabular}
    \caption{Results for I, D, S\&H and T on actual runtime data. The upper plots show RMSE and CC when training on increasing fractions of observed data during optimization and tested on unseen data from the same distribution. The lower plots show predicted values versus true observations on a log-scale for training on 100\% of the data obtained on the $QCP_{q095}$.}
    \label{fig:nnonruntime}
\end{figure*}
%
\section{Model-based Optimization}
\label{sec:eval2}
Next, we study two real-world runtime minimization problems which we first briefly describe.

\textbf{Tuning the runtime of a local-search SAT solver.} We use an existing benchmark for consistency with prior work and tune $4$ continuous hyperparameters of \saps{}~\citep{hutter-cp02a} on $3$ SAT instances as introduced by~\citet{hutter-lion10a}. The instances are quasigroup completion problems ($QCP$)~\citep{gomes-aaai97a} of increasing difficulty denoting the $50$\% ($QCP_{med}$), $75$\% ($QCP_{q075}$) and $95$\% ($QCP_{q095}$) quantiles of hardness for \saps{} \wrt a large distribution of instances. Instead of runtime in seconds, we optimize the number of steps as a more robust measurement~\citep{eggensperger-mlj18a}. We ran each optimization method for $60$min and set the overall cutoff to $10\,000$ steps. 

\textbf{Tuning the time-to-accuracy of NNs.} In the spirit of a recent effort to establish time-to-accuracy to evaluate deep learning architectures~\citep{coleman-osr19a,mattson_mlsys20} \wrt both, speed and performance, we constructed a new benchmark. We defined a $7$-dimensional configuration space consisting of commonly tuned hyperparameters of neural networks (see appendix (Table 2)) and draw $2\,000$ random configurations to evaluate their validation performance on $8$ datasets obtained from OpenML~\citep{vanschoren-sigkdd13a} using the Python-API~\citep{feurer-arxiv19a}. We selected datasets where the default configuration achieved the target accuracy in more than $5$ and less than $100$ seconds. For each dataset, we considered the $95$\% quantile of these runs as the target value to achieve for our network. As a performance measure, we use the training time in seconds to reach this preset accuracy. Furthermore, we considered runs, for which the training diverged, as timeouts with the given cutoff time, allowed each optimization method to run for $24$h and set the overall cutoff per run to $100$ seconds.

\subsection{Evaluating the quality of our NNs on runtime data}
\label{ssec:exp:nn4rtd}
\begin{table*}[t]
    \centering
    \small
\caption{Results for minimizing number of steps for \saps{} (upper) and the time-to-accuracy for NNs (lower). For each optimization method, we report the median, the lower and upper quartiles across $32$ (\saps) and $16$ (NN) repetitions. We evaluated the final configuration $1\,000$ times for \saps{} and $100$ times for the NN. We underline the best found value and boldface values that are not statistically different due to a random permutation test with $100\,000$ permutations. In the lower part of the table, we report the average rank and the averaged normalized score for each optimization method.}
    \label{tab:optresults}
\begin{tabular}{lrccc}
 Set & Default & Rand & RF w\textbackslash{} S\&H & NN w\textbackslash{} TS \& Tobit \\
\midrule
$QCP_{med}$  &    64.57 & $12.10 [11.78;12.62]$             & $12.09 [11.79;12.52]$             & $\mathbf{\underline{11.43}} [11.27;11.87]$    \\
$QCP_{q075}$ &   2889.52 & $\mathbf{27.09} [25.48;28.19]$    & $\mathbf{26.22} [25.35;27.49]$    & $\mathbf{\underline{24.99}} [24.28;25.58]$    \\
$QCP_{q095}$ &   9928.54 & $\mathbf{131.26} [125.54;146.93]$ & $\mathbf{130.33} [118.17;140.07]$ & $\mathbf{\underline{113.16}} [108.72;119.54]$ \\
\midrule
\midrule
 adult     &      8.69 & $8.69 [7.58;8.69]$            & $4.42 [4.27;5.43]$                         & $\mathbf{\underline{3.85}} [3.80;4.15]$    \\
 airlines  &     38.99 & $32.40 [28.76;36.22]$         & $\mathbf{26.82} [25.47;28.94]$             & $\mathbf{\underline{24.32}} [24.04;24.90]$ \\
 bank      &     12.17 & $11.02 [10.36;11.88]$         & $9.57 [9.24;10.10]$                        & $\mathbf{\underline{9.04}} [8.93;9.36]$    \\
 connect-4 &     47.41 & $22.55 [21.43;27.47]$         & $\mathbf{\underline{12.54}} [12.02;13.65]$ & $\mathbf{15.87} [13.97;17.27]$             \\
 credit-g  &     39.71 & $\mathbf{10.77} [5.79;19.14]$ & $\mathbf{\underline{4.52}} [4.28;5.86]$    & $\mathbf{4.98} [3.96;7.56]$                \\
 jannis    &     16.61 & $11.97 [11.30;12.41]$         & $\mathbf{10.33} [10.20;10.58]$             & $\mathbf{\underline{10.30}} [10.22;10.52]$ \\
 numerai   &     15.36 & $9.06 [8.68;9.44]$            & $\mathbf{\underline{8.37}} [8.31;8.75]$    & $\mathbf{8.67} [7.72;9.33]$                \\
 vehicle   &      7.5  & $\mathbf{3.35} [3.26;3.46]$   & $\mathbf{3.10} [2.91;3.25]$                & $\mathbf{\underline{2.80}} [2.64;3.47]$    \\
\midrule
\midrule
average rank & {-} & $3$ & $1.\overline{72}$ & $1.\overline{27}$ \\
average score & {-} & $63.77$ & $86.08$ & $91.39$ \\
\bottomrule
\end{tabular}
\end{table*}
Before turning to the optimization problem, we first study the performance of our model on data obtained from actual optimization runs. For this, we train our model posthoc on runhistories $(\conf_i, \cost_i, \ind_i)_{i=1}^n$ obtained from running random search with racing, containing all observations evaluated during optimization. We consider one task from each of our optimization problems: Tuning \saps{} on $QCP_{q095}$ and tuning time-to-accuracy on the $adult$ dataset. We used $32$ runhistories for $QCP_{q095}$ and $16$ for $adult$ and used $8$ and $4$, respectively, of these as a hold-out test set. For each configuration in the test set, we collected observations to obtain an empirical estimate of its actual performance.\footnote{To obtain results in a feasible time we applied a global cutoff yielding still some globally censored values. To obtain a ground truth value for each configuration to compare our prediction against, instead of using the empirical mean biased due to the global cutoff, we use the mean of a normal distribution fitted via maximizing the Tobit likelihood on the log-values. We note that this only makes a difference for configurations where we observed uncensored and timed-out runs. We replaced all values higher than the globally set cutoff by the cutoff before computing any metrics.}

Again, we compare I, D, S\&H, and T. We study the RMSE and the Spearman rank correlation coefficient (CC) of the predictions of individual networks (and provide more metrics in the appendix (Figure 1 and 2)), each trained for $40\ 000$ epochs and the estimated mean for each configuration. To accommodate for noise in our training procedure, we conducted $10$ repetitions, each training on all but the test runhistories.

We report the mean and standard deviation across these repetitions in Figure~\ref{fig:nnonruntime} and also additionally report the fraction of censored values present during training. The upper row shows how RMSE and CC develop using more training data by training on increasing fractions of the runhistories (i.e. studying the model's performance at different time steps of the optimization run). We observe that for both scenarios the Tobit loss clearly outperforms alternative handling of censored information. Furthermore, for both scenarios RMSE values for T stay the same ($QCP_{q095}$) or decrease (\emph{adult}) with more data while the values for other methods strongly increase ($QCP_{q095}$) or stagnate (\emph{adult}), highlighting the benefit obtained from using our loss function. The CC values indicate how well a method preserves the ranking between configurations and again, T performs best followed by S\&H. Both, D and I, perform worse due to underestimating the true values.
In the second row, we plot actual predictions against the empirical mean showing the difference in the quality of the predictions. While T preserves the ranking between configurations best, D and I tend to predict the training data mean, making these bad choices for the optimization procedure. S\&H performs slightly better but does not restore censored values with the same quality as T.

\subsection{NNs for Algorithm Configuration}
\label{ssec:exp:nn4ac}

Now, we turn to the results of using our model on $11$ minimization tasks. As a competitive baseline, we use the state-of-the-art sequential model-based optimization framework SMAC~\citep{hutter-lion11a} using RFs to handle censored data with S\&H~\cite{hutter-bayesopt11}, which we replace with our NNs using Thompson Sampling (NN w\textbackslash{} TS \& Tobit).
Additionally, we ran random search (\roar{}) as a baseline to quantify the contribution of a model to the optimization procedure. We ran the experiments as discussed above and report the median performance of all methods in Table~\ref{tab:optresults}. Additionally, we report the average rank and the average normalized score across all scenarios (see appendix (Table 3) for more details).

In general, all methods found configurations that substantially improve over the default configuration being more than $80\times$ (\saps{} on $QCP_{q095}$) and $8\times$ (time-to-accuracy on credit-g) faster. Furthermore, all model-based methods perform better than \roar{} for all benchmarks and almost always statistically significantly so.
Looking at the \saps{} results and comparing the RF-based method to our method, we found that NN w\textbackslash{} TS \& Tobit performed better on all scenarios indicating that the networks with Thompson sampling indeed work well for such low-dimensional and noisy benchmarks. On the higher dimensional tasks minimizing time-to-accuracy, the RF model performs more competitively, but still NN w\textbackslash{} TS \& Tobit performs significantly better on $2$ out of $8$ tasks and better or competitively on the remaining datasets. Also, looking at the aggregated rank and score, NN w\textbackslash{} TS \& Tobit obtained the lowest rank and the highest score showing that NNs are a promising alternative.

\section{Conclusion and Future Work}
\label{sec:conclusion}
Better methods for efficiently handling censored data directly lead to improvements for model-based optimization in a wide variety of domains, e.g., minimizing the time-to-accuracy for machine learning algorithms and the solution time for combinatorial optimization. Recently, the potential of NNs to rival and outperform GP- and RF-based Bayesian optimization has been demonstrated, but they have not yet been extended for optimization procedures relying on censoring strategies as one of their primary factor for efficiency. In this work, we propose a theoretically motivated loss function~\citep{tobin-econometria58} which directly incorporates censored data, and thus, the missing piece for NN-driven optimization procedures for these domains. We empirically showed how well our loss function works for censored data and evaluated our solution on real optimization benchmarks outperforming the previous state of the art.

Building on these promising results, for future work, it would be interesting to investigate how to extend our model for training on non-Gaussian distributions and in the joint space of configurations and tasks~\cite{eggensperger-ijcai18a}. Furthermore, another interesting direction would be to study whether the model capacity of the NNs can be adapted over time during the optimization~\citep{franke-meta19new}. Finally, we believe the overhead of training the model can be further reduced by studying how to reuse trained NNs during optimization~\cite{huang-iclr17a}.

\newpage
\paragraph{Acknowledgements}
This work has partly been supported by the European Research Council (ERC) under the European Union’s  Horizon 2020 research and innovation programme under grant no. 716721. Robert Bosch GmbH is  acknowledged for financial support. 

\bibliography{shortstrings,lib,local,shortproc}

\includepdf[pages=1-]{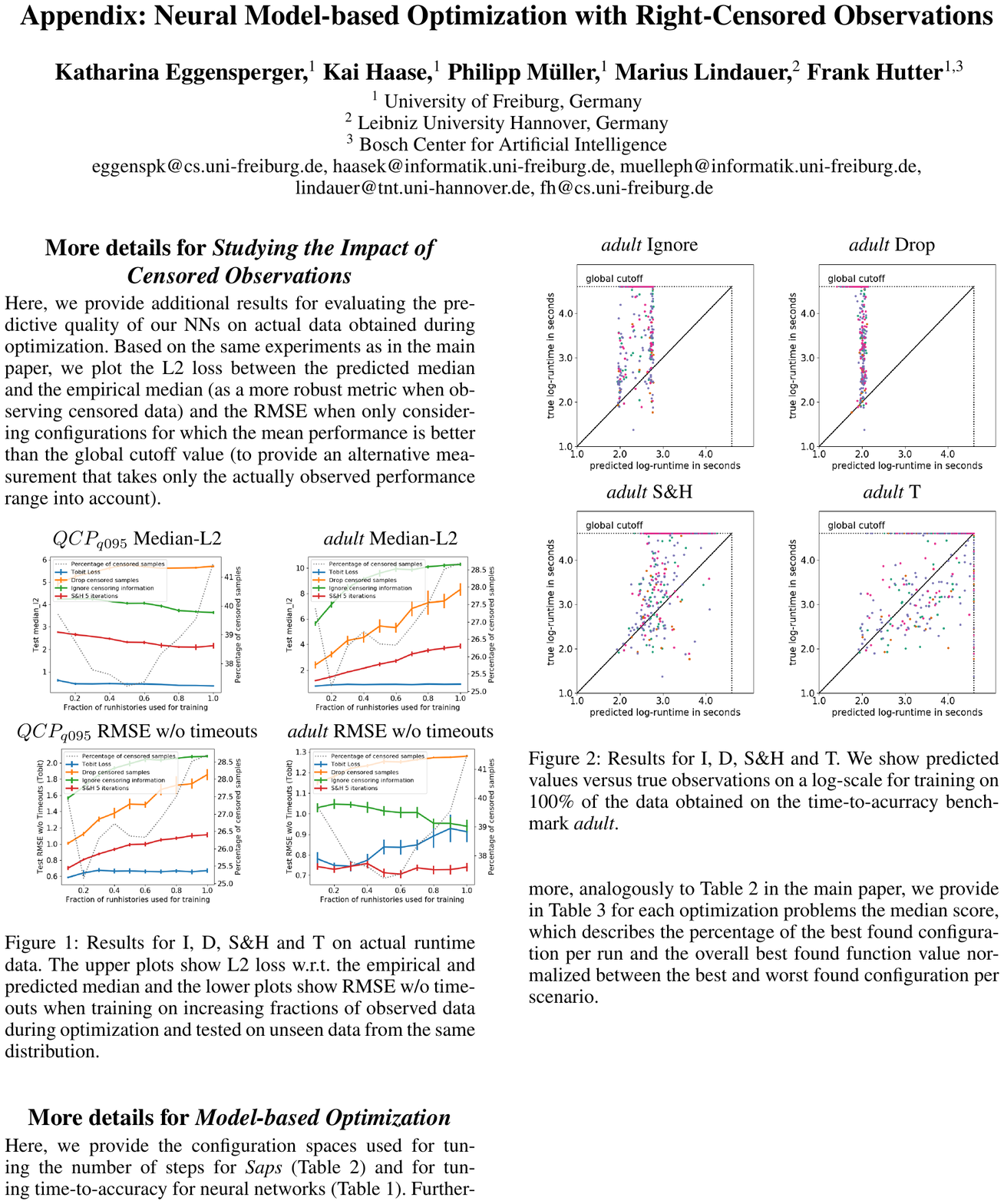}

\end{document}